\documentclass[10pt,journal,compsoc]{IEEEtran}
\usepackage{epsfig}
\usepackage{graphicx}
\usepackage{amsmath}
\usepackage{amssymb}
\usepackage{multirow}
\usepackage{makecell}
\usepackage{tabulary}
\usepackage{listofitems}
\usepackage{stackengine}

\usepackage{url}
\usepackage[pagebackref=true,breaklinks=true,letterpaper=true,colorlinks,bookmarks=false]{hyperref}

\ifCLASSOPTIONcompsoc
  \usepackage[nocompress]{cite}
\else
  \usepackage{cite}
\fi
\ifCLASSINFOpdf
\else
\fi
\begin{document}
%
\title{HIRI-ViT: Scaling Vision Transformer\\ with High Resolution Inputs}
%
%
%
%

\author{
Ting~Yao,~\IEEEmembership{Senior Member,~IEEE},
Yehao~Li,
Yingwei~Pan,~\IEEEmembership{Member,~IEEE},
and~Tao~Mei,~\IEEEmembership{Fellow,~IEEE}
\IEEEcompsocitemizethanks{\IEEEcompsocthanksitem
Ting~Yao, Yehao~Li, Yingwei~Pan, and Tao~Mei are with HiDream.ai. (e-mail: tingyao.ustc@gmail.com; yehaoli.sysu@gmail.com; panyw.ustc@gmail.com; tmei@live.com). Yingwei Pan is the corresponding author.}
}

%
%

\markboth{IEEE TRANSACTIONS ON PATTERN ANALYSIS AND MACHINE INTELLIGENCE}%
{Yao \MakeLowercase{\textit{et al.}}: HIRI-ViT: Scaling Vision Transformer with High Resolution Inputs}
%



\IEEEtitleabstractindextext{%
\begin{abstract}
The hybrid deep models of Vision Transformer (ViT) and Convolution Neural Network (CNN) have emerged as a powerful class of backbones for vision tasks. Scaling up the input resolution of such hybrid backbones naturally strengthes model capacity, but inevitably suffers from heavy computational cost that scales quadratically. Instead, we present a new hybrid backbone with HIgh-Resolution Inputs (namely HIRI-ViT), that upgrades prevalent four-stage ViT to five-stage ViT tailored for high-resolution inputs. HIRI-ViT is built upon the seminal idea of decomposing the typical CNN operations into two parallel CNN branches in a cost-efficient manner. One high-resolution branch directly takes primary high-resolution features as inputs, but uses less convolution operations. The other low-resolution branch first performs down-sampling and then utilizes more convolution operations over such low-resolution features. Experiments on both recognition task (ImageNet-1K dataset) and dense prediction tasks (COCO and ADE20K datasets) demonstrate the superiority of HIRI-ViT. More remarkably, under comparable computational cost ($\sim$5.0 GFLOPs), HIRI-ViT achieves to-date the best published Top-1 accuracy of 84.3\% on ImageNet with 448$\times$448 inputs, which absolutely improves 83.4\% of iFormer-S by 0.9\% with 224$\times$224 inputs.
\end{abstract}

}

\maketitle

\IEEEdisplaynontitleabstractindextext

%
\IEEEpeerreviewmaketitle

\IEEEraisesectionheading{\section{Introduction}\label{sec:introduction}}

\IEEEPARstart{I}{nspired} by the dominant Transformer structure in Natural Language Processing (NLP) \cite{vaswani2017attention}, the Computer Vision (CV) field witnesses the rise of Vision Transformer (ViT) for designing vision backbones. This trend has been most visible for image/action recognition \cite{dosovitskiy2020image,wavevit2022,long2022stand,long2022dtf} and dense prediction tasks like object detection \cite{carion2020end}. Many of these success can be attributed to the flexible modeling of long-range interaction among input visual tokens via self-attention mechanism in conventional Transformer block. Most recently, several concurrent studies \cite{d2021convit,guo2022cmt,wu2021cvt,xiao2021early,yuan2021incorporating} point out that it is sub-optimal to directly employ pure Transformer block over visual token sequence. Such design inevitably lacks the right inductive bias of 2D regional structure modeling. To alleviate this limitation, they lead the new wave of instilling the 2D inductive bias of Convolution Neural Network (CNN) into ViT, yielding the CNN+ViT hybrid backbones.

\begin{figure}[!tb]
\vspace{-0.0in}
\centering {\includegraphics[width=0.5\textwidth]{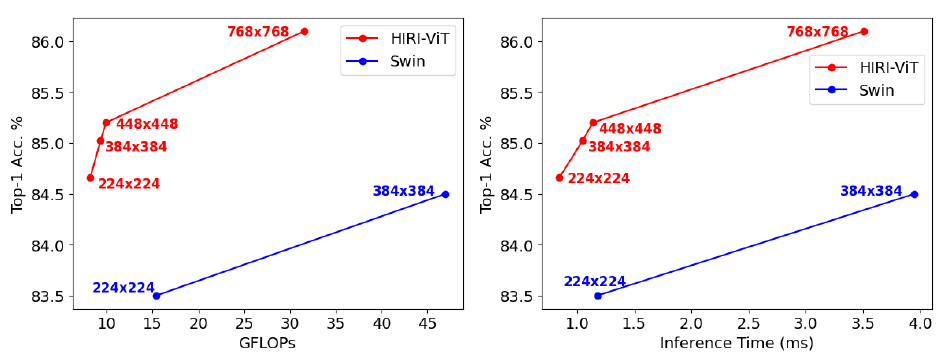}}
\vspace{-0.22in}
\caption{\small Performance and computational cost (i.e., GFLOPs and Inference time) of Swin Transformer and our HIRI-ViT with different input resolutions (224$\times$224, 384$\times$384, and 448$\times$448) on ImageNet-1K dataset.}
\label{fig:intro}
\end{figure}

A common practice in CNN backbone design is to enlarge network depth/width/input resolution \cite{huang2019gpipe,tan2019efficientnet}, thereby enhancing model capacity by capturing more fine-grained patterns within inputs. Similar in spirit, our work wants to delve into the process of scaling CNN+ViT hybrid backbones with high resolution inputs. Nevertheless, in analogy to the scaling of CNN backbones, simply enlarging the input resolution of prevalent ViT backbones will introduce practical challenge especially the sharply increased computational costs. Taking a widely adopted ViT backbone, Swin Transformer \cite{liu2021swin}, as an example, when directly enlarging the input resolution from 224$\times$224 to 384$\times$384, the top-1 accuracy on ImageNet-1K is evidently increased from 83.5\% to 84.5\%. However, as shown in Figure \ref{fig:intro}, the computational cost of Swin Transformer with 384$\times$384 inputs (GFLOPs: 47.0, Inference time: 3.95 ms) is significantly heavier than that with 224$\times$224 inputs (GFLOPs: 15.4, Inference time: 1.17 ms).

In view of this issue, our central question is – is there a principled way to scale up CNN+ViT hybrid backbones with high resolution inputs while maintaining comparable computational overhead?
To this end, we devise a family of five-stage Vision Transformers tailored for high-resolution inputs, which contain two-branch building blocks in earlier stages that seek better balance between performance and computational cost. Specifically, the key component in remoulded stem/CNN block is a combination of high-resolution branch (with less convolution operations over high-resolution inputs) and low-resolution branch (more convolution operations over low-resolution inputs) in a parallel. Such two-branch design takes the place of a single branch with standard convolution operations in stem/CNN blocks. By doing so, we not only preserve the strengthened model capacity with high-resolution inputs, but also significantly reduce the computational cost with light-weight design of each branch. As shown in Figure \ref{fig:intro}, by enlarging the input resolution from 224$\times$224 to 384$\times$384, a clear performance boost is attained for our HIRI-ViT, but the computational cost only slightly increases (GFLOPs: 8.2 to 9.3, Inference time: 0.84 ms to 1.04 ms). Even when enlarging the input resolution to 768$\times$768, HIRI-ViT leads to significant performance improvements, while requiring less computational cost than Swin Transformer.

By integrating this two-branch design into CNN+ViT hybrid backbone, we present a new principled five-stage vision backbone, namely HIRI-ViT, that efficiently scales up Vision Transformer with high-resolution inputs. In particular, we first upgrade the typical Conv-stem block by decomposing the single CNN branch into two parallel branches (i.e., high-resolution and low-resolution branches), leading to high resolution stem (HR-stem) blocks. Next, CNN blocks in earlier stages are remoulded by replacing CNN branch with the proposed two-branch design. Such new High Resolution block (HR-block) triggers a cost-efficient encoding of high-resolution inputs.

The main contribution of this work is the proposal of scaling up the CNN+ViT hybrid backbone with high resolution inputs while maintaining favorable computational cost. This also leads to the elegant design of decomposing the typical CNN operations over high resolution inputs into two parallel light-weight CNN branches. Through an extensive set of experiments over multiple vision tasks (e.g., image recognition, object detection and instance/semantic segmentation), we demonstrate the superiority of our new HIRI-ViT backbone in comparison to the state-of-the-art ViT and CNN backbones under comparable computational~cost.

\begin{figure*}[!tb]
\vspace{-0.0in}
\centering {\includegraphics[width=1\textwidth]{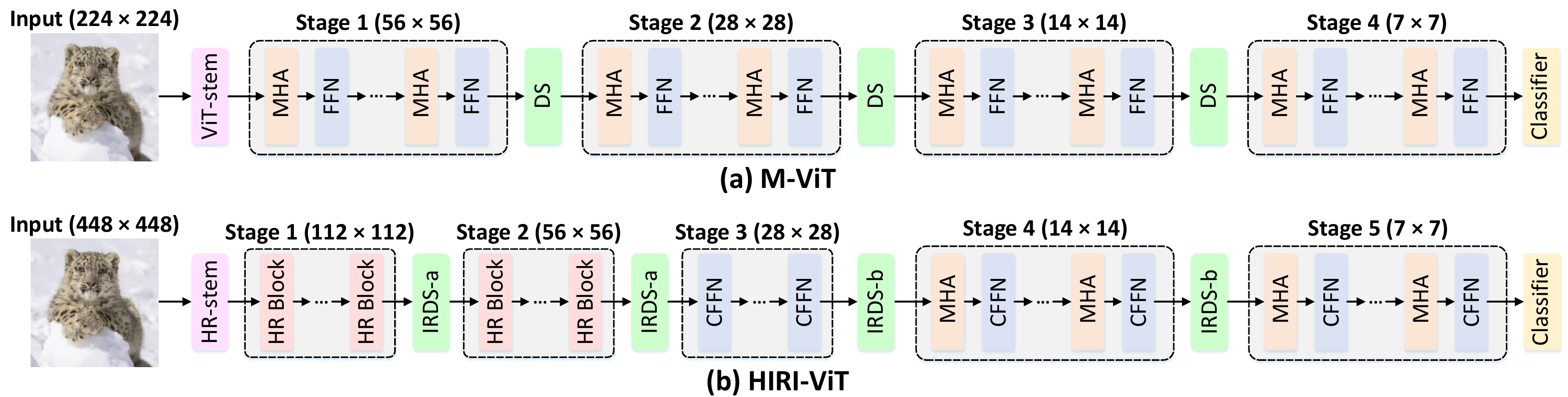}}
\caption{\small Comparison between (a) the typical multi-stage Vision Transformer (M-ViT) with regular resolution inputs and (b) our HIRI-ViT with high resolution inputs. HR-stem: High Resolution Stem.  IRDS: Inverted Residual Downsampling. HR Block: High Resolution Block. CFFN: Convolutional Feed-Forward Network.}
\label{fig:net}
\end{figure*}

\section{Related Work}
\subsection{Convolutional Neural Network}
Inspired by the breakthrough via AlexNet \cite{krizhevsky2012imagenet} on ImageNet-1K benchmark, Convolutional Neural Networks (CNN) have become the de-facto backbones in computer vision field. Specifically, one of the pioneering works is VGG \cite{simonyan2014very}, which increases the network depth to enhance the model capability. ResNet \cite{he2016deep} trains deeper networks by introducing skip connections between convolutional blocks, leading to better generalization and impressive results. DenseNet \cite{huang2017densely} further scales up the network with hundreds of layers by connecting each convolutional block to all the previous blocks. Besides going deeper, designing multi-branch block is another direction to enhance model capacity. InceptionNet \cite{szegedy2015going,szegedy2016rethinking} integrates multiple paths with different kernels into a single convolutional block via the split-transform-merge strategy. ResNeXt \cite{xie2017aggregated} demonstrates that increasing cardinality with the homogeneous and multi-branch architecture is an effective way of improving performance. Res2Net \cite{gao2019res2net} develops multiple receptive fields at a more granular level by constructing hierarchical residual-like connections. EfficientNet \cite{tan2019efficientnet} exploits neural architecture search to seek a better balance between network width, depth, and resolution. Recently, ConvNeXt \cite{liu2022convnet,woo2023convnext} modernizes ResNet by integrating it with Transformer designs, achieving competitive results gainst Vision Transformer while retaining the efficiency of CNN.

\subsection{Vision Transformer}
Inspired by Transformer \cite{vaswani2017attention} in NLP field, Vision Transformer architectures start to dominate the construction of backbones in vision tasks recently. The debut of Vision Transformer \cite{dosovitskiy2020image} splits the image into a sequence of patches (i.e., visual tokens) and then directly applies self-attention over the visual tokens. DeiT \cite{touvron2021training} learns Vision Transformer in a data-efficient manner with the upgraded training strategies and distillation procedure. Since all layers of ViT/DeiT are designed under the same lower resolution, it might not be suitable to directly apply them for dense prediction tasks \cite{wang2021pyramid}. To address this issue, PVT \cite{wang2021pyramid} adopts a pyramid structure of ViT with four stages, whose resolutions progressively shrink from high to low. Swin \cite{liu2021swin} integrates shifted windowing scheme into local self-attention, allowing cross-window connection with linear computational complexity. Twins \cite{chu2021twins} interleaves locally-grouped attention and global sub-sampled attention to exploit both fine-grained and long-distance global information. DaViT \cite{ding2022davit} further proposes dual attention mechanism with spatial window attention and channel group attention, aiming to enable local fine-grained and global interactions. Later on, CNN and ViT start to interact with each other, yielding numerous hybrid backbones. In particular, CvT \cite{wu2021cvt} and CeiT \cite{yuan2021incorporating} upgrade self-attention and feed-forward module with convolutions respectively. ViTAE \cite{zhang2022vitaev2} introduces an extra convolution block in parallel to self-attention module, whose outputs are fused and fed into feed-forward module. iFormer \cite{siinception} couples max-pooling, convolution, and self-attention to learn both high- and low-frequency information. MaxViT \cite{tu2022maxvit} performs both local and global spatial interactions by combining convolution, local self-attention and dilated global self-attention into a single block.

\subsection{High-resolution Representation Learning}
Significant advancements have been made in exploring high-resolution inputs in CNN backbone design. For example, HRNet \cite{wang2020deep} maintains the high-resolution branch through the whole network, and fuses multiresolution features repeatedly. EfficientHRNet \cite{neff2021efficienthrnet} further unifies EfficientNet and HRNet, and designs a downwards scaling method to scale down the input resolution, backbone network, and high-resolution feature network. Later on, Lite-HRNet \cite{yu2021lite} enhances the efficiency of HRNet by applying shuffle blocks and conditional channel weighting unit. Subsequently, several works start to build Transformer backbones with high-resolution inputs. In between, HR-NAS \cite{ding2021hr} introduce a multi-resolution search space including both CNN and Transformer blocks for multi-scale information and global context modeling. Recently, HRViT \cite{gu2022multi} and HRFormer \cite{YuanFHLZCW21} target for constructing Vision Transformer with multi-scale inputs by keeping all resolutions throughout the network and performing cross-resolution interaction. Nevertheless, the inputs of those ViT backbones are still limited to a small resolution (i.e., 224$\times$224). Even though most hybrid backbones can be directly scaled up with higher resolution (e.g., 384$\times$384), the computation cost becomes much heavier, which scales quadratically w.r.t. the input resolution. Instead, our work paves a new way to scale up the CNN+ViT hybrid backbone with high resolution inputs, and meanwhile preserve the favorable computational overhead as in small resolution.

\section{Preliminaries}
Conventional multi-stage Vision Transformer (M-ViT) \cite{chen2021regionvit,chen2021visformer,liu2021swin,wang2021pvtv2} is commonly composed by a stem plus four stages as in ConvNets \cite{he2016deep,liu2022convnet,xie2017aggregated} (see Figure \ref{fig:net} (a)). Specifically, the stem layer is first utilized to split the input image (resolution: 224$\times$224) into patches. Each patch is regarded as a ``visual token'' and will be further fed into the following stages. Each stage contains multiple Transformer blocks, and each Transformer block consists of a multi-head self-attention module ($\bf{MHA}$) followed by a feed-forward network ($\bf{FFN}$). Typically, a downsampling layer ($\bf{DS}$) is inserted between every two stages to merge the input ``visual tokens'' (i.e., reduce the resolution of the feature map) and meanwhile enlarge their channel dimension. Finally, a classifier layer is adopted to predict the probability distribution based on the last feature map.

\textbf{Multi-head Self-attention.} Multi-head self-attention targets for capturing long-range dependencies among the visual tokens. Technically, let $X \in {{\mathbb{R}}^{n \times D}}$ denote the features of visual tokens, where $n = H \times W$ is the number of visual tokens and $H$/$W$/$D$ represents the height/width/channel number respectively. The input $X$ is first linearly transformed into queries $Q \in {{\mathbb{R}}^{n \times D}}$, keys $K \in {{\mathbb{R}}^{n \times D}}$, and values $V \in {{\mathbb{R}}^{n \times D}}$, which are further decomposed into $N_h$ heads/parts along channel dimension. By denoting the queries, keys, and values of the $j$-th head as ${Q_j} \in {{\mathbb{R}}^{n \times {D_h}}}$, ${K_j} \in {{\mathbb{R}}^{n \times {D_h}}}$, and ${V_j} \in {{\mathbb{R}}^{n \times {D_h}}}$ ($D_h$: the dimension of each head), the self-attention module operates as follows:
\begin{equation}\small
\label{eq:sa1}
\begin{aligned}
&{{\bf{MultiHead}}(Q,K,V) = {\bf{Concat}}(head_0,...,head_{N_h})W^O},\\
&{head_j = {\bf{Attention}}(Q_j,K_j,V_j)},\\
&{{\bf{Attention}}(Q_j,K_j,V_j) = {\bf{Softmax}}(\frac{{{Q_j}{{K_j}^T}}}{{\sqrt {D_h} }}){V_j}},
\end{aligned}
\end{equation}
where $W^O$ is weight matrix and ${\bf{Concat}}(\cdot)$ is the concatenation operation. Considering that the computational cost of self-attention scales quadratically w.r.t. the token number, spatial reduction is usually applied over keys/values to reduce the computational/memory overhead \cite{guo2022cmt,wang2021pvtv2}.

\textbf{Feed-Forward Network.} The original feed-forward network \cite{dosovitskiy2020image,vaswani2017attention} consists of two fully-connected ($\bf{FC}$) layers coupled with a non-linear activation in between:
\begin{equation}\small
\label{eq:ffn}
{\bf{FFN}}(X) = {\bf{FC}}(\sigma ({\bf{FC}}(X))),
\end{equation}
where $\sigma$ denotes the non-linear activation. Inspired by \cite{guo2022cmt,wang2021pvtv2}, we upgrade $\bf{FFN}$ with an additional convolutional operation to impose 2D inductive bias, yielding Convolutional Feed-Forward Network ($\bf{CFFN}$). The overall operations of this $\bf{CFFN}$ are summarized as
\begin{equation}\small
\label{eq:cffn}
{\bf{CFFN}}(X) = {\bf{FC}}({\bf{DWConv}}(z) + z),~z = \sigma ({\bf{FC}}(X)),
\end{equation}
where $\bf{DWConv}$ denotes the depth-wise convolution.

\section{HIRI-ViT}
In this paper, our goal is to design a principled Transformer structure (namely HIRI-ViT) that enables a cost-efficient scaling up of Vision Transformer with high resolution inputs. To do so, we upgrade typical four-stage M-ViT into a new family of five-stage ViT that contain two-branch building blocks in earlier stage, which decompose the single-branch CNN operations into two parallel CNN branches. This way leads to favorable computational overhead tailored for high resolution inputs. Figure \ref{fig:net} (b) illustrates an overall architecture of our HIRI-ViT.

\subsection{High Resolution Stem}
\begin{figure}[!tb]
\centering {\includegraphics[width=0.5\textwidth]{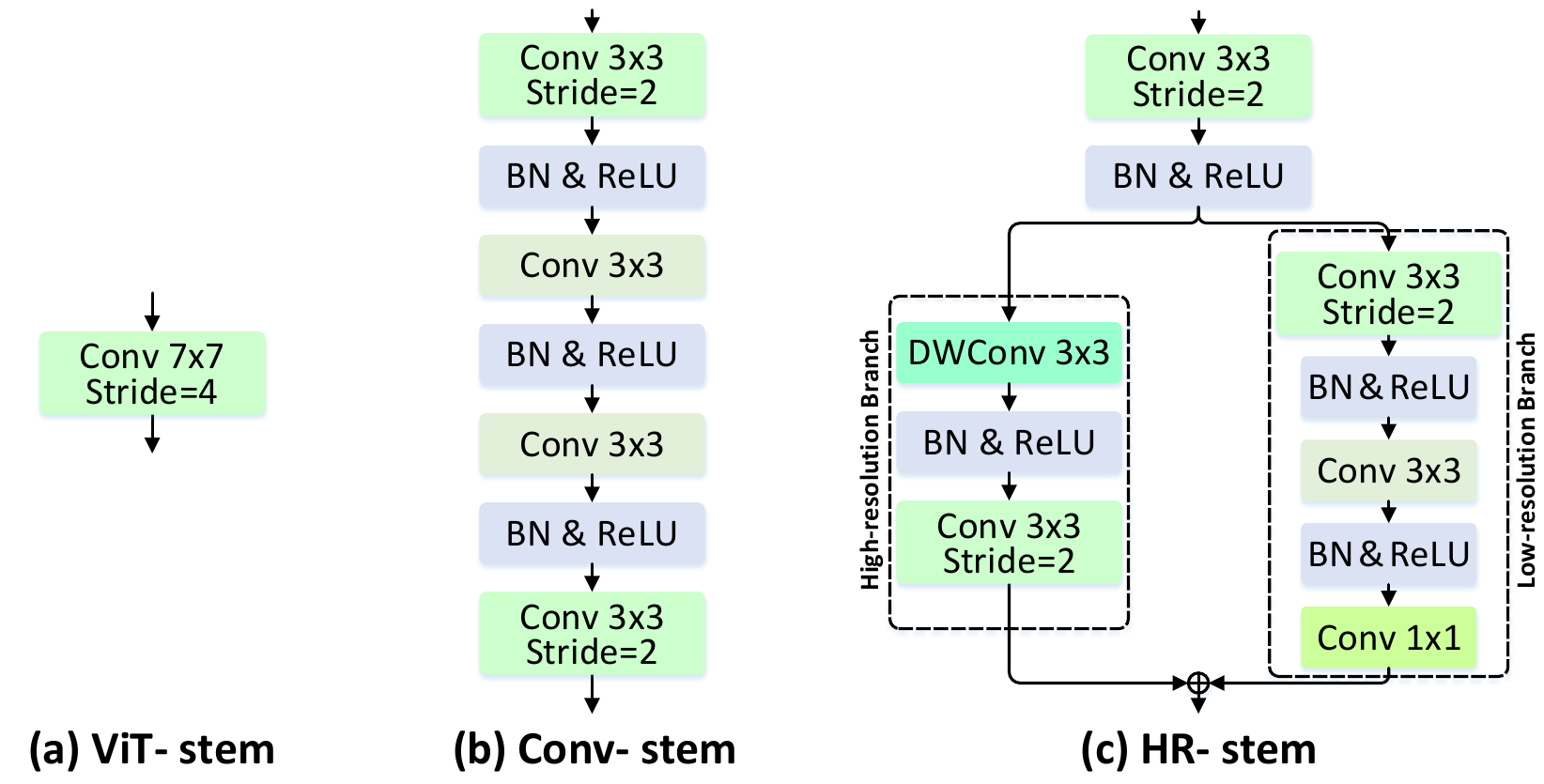}}
\caption{An illustration of (a) ViT-stem and (b) Conv-stem in typical Vision Transformer, and (c) our HR-stem in HIRI-ViT.}
\label{fig:stem}
\end{figure}

The design of stem layer in conventional Vision Transformer can be briefly grouped into two dimensions: ViT-stem and Conv-stem \cite{he2019bag,wang2022scaled}. As shown in Figure \ref{fig:stem} (a), ViT-stem is implemented as a single strided convolution layer (e.g., stride = 4, kernel size = 7 \cite{wang2021pvtv2}) which aims to divide the input image into patches. Recently, \cite{xiao2021early,wang2022scaled} reveals that replacing the ViT-stem with several stacked 3$\times$3 convolutions (i.e., Conv-stem shown in Figure \ref{fig:stem} (b)) can stabilize the network optimization procedure and meanwhile improve peak performance. Conv-stem results in a slight increase of the parameters and GFLOPs for typical input resolution (e.g., 224$\times$224). However, when the input resolution significantly increases (e.g., 448$\times$448), the GFLOPs of Conv-stem become much heavier than ViT-stem. To alleviate these issues, we design a new High Resolution stem layer (HR-stem in Figure \ref{fig:stem} (c)) by remoulding the single-branch Conv-stem as two parallel CNN branches. Such design not only preserves high model capacity as Conv-stem, but also consumes favorable computational cost under high resolution inputs.

Technically, HR-stem first utilizes a strided convolution (stride = 2, kernel size = 3) to downsample the input image as in Conv-stem. After that, the downsampled feature map is fed into two parallel branches (i.e., high and low-resolution branches). The high-resolution branch contains a light-weight depth-wise convolution followed by a strided convolution. For the low-resolution branch, a strided convolution is first employed to downsample the feature map. Then two convolutions (3$\times$3 and 1$\times$1 convolutions) are applied to impose inductive bias. Finally, the output of HR-stem is achieved by aggregating the two branches and the sum is further normalized via batch normalization ($\bf{BN}$).

\subsection{High Resolution Block}
\begin{figure}[!tb]
\centering {\includegraphics[width=0.5\textwidth]{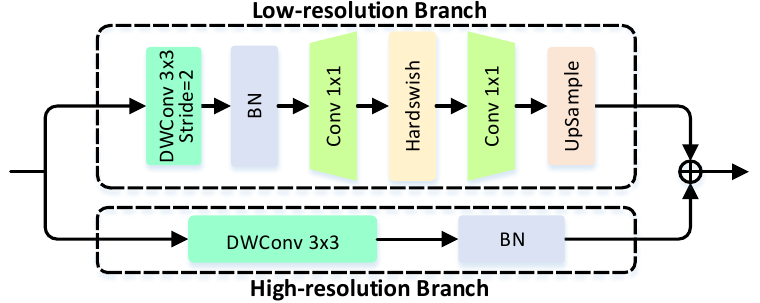}}
\caption{An illustration of HR block in HIRI-ViT.}
\label{fig:hrblock}
\end{figure}

In view of the fact that the input resolution of the first two stages in hybrid backbones is large, the computational costs are relatively higher for Transformer blocks. To address this limitation, we replace the Transformer blocks in the first two stages with our new High resolution block (HR block), enabling a cost-efficient encoding of high-resolution inputs in the earlier stages. Specifically, similar to HR-stem, HR block is composed of two branches in parallel. The light-weight high-resolution branch captures coarse-level information over high-resolution inputs, while the low-resolution branch utilizes more convolution operations to extract high-level semantics over low-resolution inputs.
Figure \ref{fig:hrblock} depicts the detailed architecture of HR block.

Concretely, the high-resolution branch is implemented as a light-weight depth-wise convolution. For low-resolution branch, a strided depth-wise convolution (stride = 2, kernel size = 3) with $\bf{BN}$ is first utilized to downsample the input feature map. Then, a feed-forward operation (i.e., two $\bf{FC}$ with an activation in between) is applied over the low-resolution feature map. After that, the low-resolution output is upsampled via repetition, which is further fused with the high-resolution output.

\subsection{Inverted Residual Downsampling}
\begin{figure}[!tb]
\centering {\includegraphics[width=0.5\textwidth]{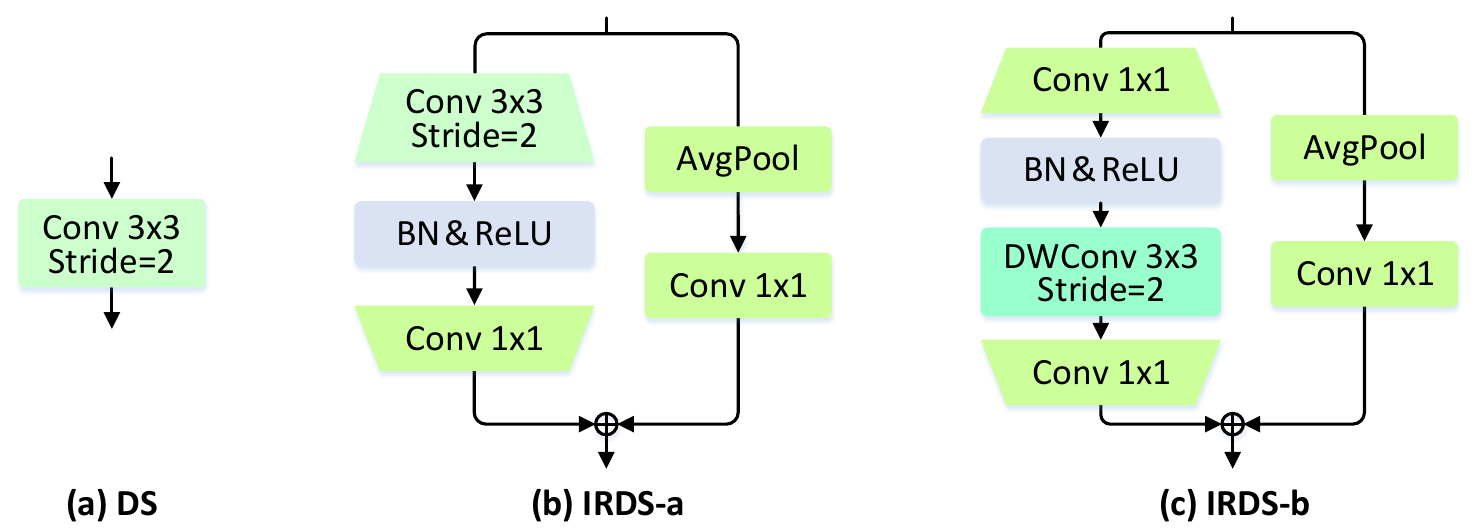}}
\caption{An illustration of (a) downsampling (DS) in typical Vision Transformer and two variants of our Inverted Residual Downsampling (i.e., (b) IRDS-a and (c) IRDS-b) in HIRI-ViT.}
\label{fig:ds}
\end{figure}

In conventional M-ViT, the spatial downsampling is performed through a single strided convolution (e.g., stride = 2, kernel size = 3 \cite{wang2021pvtv2}), as shown in Figure \ref{fig:ds} (a). Inspired by ConvNets \cite{he2016deep,tan2021efficientnetv2}, we design a more powerful downsampling layer with two parallel branches, namely Inverted Residual Downsampling (IRDS). In particular, for the first two stages with high resolution inputs, we adopt IRDS-a (Figure \ref{fig:ds} (b)) for downsampling. IRDS-a first uses a strided $3\times3$ convolution to expand the dimension and reduce the spatial size, and then a $1\times1$ convolution is utilized to shrink the channel dimension. For the last two downsampling layers, we leverage IRDS-b (Figure \ref{fig:ds} (c)), which is similar to inverted residual block \cite{sandler2018mobilenetv2}. The difference lies in that we only apply normalization and activation operations after the first convolution. Note that we add extra downsampling shortcuts to stabilize the training procedure.

\begin{table*}[!tb]\small
\centering
\caption{Model architecture specifications for three variants of our HIRI-ViT in different model sizes, i.e., HIRI-ViT-S (Small size), HIRI-ViT-B (Base size), and HIRI-ViT-L (Large size). $H_i$, $C_i$, and $E_i$ represents the head number, channel dimension, and expansion ratio of feed-forward layer in the stage $i$, respectively.}
\begin{tabular}{ccccc}
\Xhline{2\arrayrulewidth}
Specification & & HIRI-ViT-S & HIRI-ViT-B & HIRI-ViT-L \\ \hline
High Resolution Stem & & 4 $\times$ 4 & 4 $\times$ 4 & 4 $\times$ 4 \\ \hline
Stage 1 ($\frac{H}{4} \times \frac{W}{4}$)~~~~
        & $\begin{array}{c} {\bf{High~Resolution}} \\ {\bf{Block}} \end{array}$
        & $\left[ \begin{array}{c}  C_1=32 \\ E_1=4  \end{array} \right] \!\times\! 2$ ~
        & $\left[ \begin{array}{c}  C_1=64 \\ E_1=4  \end{array} \right] \!\times\! 2$ ~
        & $\left[ \begin{array}{c}  C_1=80 \\ E_1=4  \end{array} \right] \!\times\! 4$
        \\ \hline
Inverted Residual Downsampling-a & & 2 $\times$ 2 & 2 $\times$ 2 & 2 $\times$ 2 \\ \hline
Stage 2 ($\frac{H}{8} \times \frac{W}{8}$)~~~~
        & $\begin{array}{c} {\bf{High~Resolution}} \\ {\bf{Block}} \end{array}$
        & $\left[ \begin{array}{c}  C_2=64 \\ E_2=4  \end{array} \right] \!\times\! 2$ ~
        & $\left[ \begin{array}{c}  C_2=96 \\ E_2=4  \end{array} \right] \!\times\! 2$ ~
        & $\left[ \begin{array}{c}  C_2=160 \\ E_2=4 \end{array} \right] \!\times\! 4$
        \\ \hline
Inverted Residual Downsampling-a & & 2 $\times$ 2 & 2 $\times$ 2 & 2 $\times$ 2 \\ \hline
Stage 3 ($\frac{H}{16} \times \frac{W}{16}$)~~~~
        & $\begin{array}{c} {\bf{CFFN}} \\ {\bf{Block}} \end{array}$
        & $\left[ \begin{array}{c}  C_3=128 \\ E_3=6 \end{array} \right] \!\times\! 2$ ~
        & $\left[ \begin{array}{c}  C_3=192 \\ E_3=5 \end{array} \right] \!\times\! 3$ ~
        & $\left[ \begin{array}{c}  C_3=224 \\ E_3=5 \end{array} \right] \!\times\! 5$
        \\ \hline
Inverted Residual Downsampling-b & & 2 $\times$ 2 & 2 $\times$ 2 & 2 $\times$ 2 \\ \hline
Stage 4 ($\frac{H}{32} \times \frac{W}{32}$)~~~~
        & $\begin{array}{c} {\bf{Transformer}} \\ {\bf{Block}} \end{array}$
        & $\left[ \begin{array}{c}  H_4=5 \\ C_4=320 \\ E_4=5   \end{array} \right] \!\times\! 9$ ~
        & $\left[ \begin{array}{c}  H_4=5 \\ C_4=320 \\ E_4=4  \end{array} \right] \!\times\! 17$ ~
        & $\left[ \begin{array}{c}  H_4=7 \\ C_4=448 \\ E_4=3  \end{array} \right] \!\times\! 25$
        \\ \hline
Inverted Residual Downsampling-b & & 2 $\times$ 2 & 2 $\times$ 2 & 2 $\times$ 2 \\ \hline
Stage 5 ($\frac{H}{64} \times \frac{W}{64}$)~~~~
        & $\begin{array}{c} {\bf{Transformer}} \\ {\bf{Block}} \end{array}$
        & $\left[ \begin{array}{c}  H_5=8 \\ C_5=512 \\ E_5=5 \end{array} \right] \!\times\! 4$ ~
        & $\left[ \begin{array}{c}  H_5=10 \\ C_5=640 \\ E_5=5 \end{array} \right] \!\times\! 4$ ~
        & $\left[ \begin{array}{c}  H_5=10 \\ C_5=640 \\ E_5=5 \end{array} \right] \!\times\! 5$
        \\ \Xhline{2\arrayrulewidth}
\end{tabular}
\label{table:architecture}
\end{table*}

\begin{table}[!tb]\scriptsize
\centering
\caption{Performance comparison on ImageNet-1K validation set and ImageNet V2 matched frequency test set (V2 Top-1). We group the backbones as in \cite{chen2022mobile}. Standard deviation ($\pm x$) is reported with 10 different runs.}
\vspace{-0.1in}
\setlength{\tabcolsep}{0.0pt}
\begin{tabular}{c|c|c|c|c|cc|c}
\Xhline{2\arrayrulewidth}
Network   & Res.    & Params & GFLOPs & Throughput & Top-1 & Top-5 & V2 Top-1 \\ \hline
ResNet-50 \cite{he2016deep}             & 224  & 25.5M  & 4.1  &  3947   &  79.8  &  -    & -     \\
HRNet-W18 \cite{wang2020deep}           & 224  & 21.3M  & 4.3  &  627    & 79.9   & 94.9  & 68.4  \\
HRFormer-S \cite{YuanFHLZCW21}          & 224  & 13.5M  & 3.6  &  751    & 81.2   & 95.6  & 70.0  \\
Swin-T \cite{liu2021swin}               & 224  & 29.0M  & 4.5  &  1637   & 81.2   & 95.5  & 69.7  \\
DeiT-S \cite{touvron2021training}       & 224  & 22.1M  & 4.6  &  2245   & 81.4   &  -    & -     \\
CvT-13 \cite{wu2021cvt}                 & 224  & 20.0M  & 4.5  &  1954   & 81.6   & -     & -     \\
SE-CoTNetD-50  \cite{li2022contextual}  & 224  & 23.1M  & 4.1  &  2687   & 81.6   & 95.8  & 71.5  \\
CeiT-S \cite{yuan2021incorporating}     & 224  & 24.2M  & 4.8  &  1537   & 82.0   & 95.9  & -     \\
PVTv2-B2   \cite{wang2021pvtv2}         & 224  & 25.4M  & 4.0  &  1296   & 82.0   & 95.6  & 71.6  \\
ConvNeXt-T \cite{liu2022convnet}        & 224  & 28.6M  & 4.5  &  1577   & 82.1   & 95.9  & 71.0  \\
Visformer-S \cite{chen2021visformer}    & 224  & 40.2M  & 4.9  &  2128   & 82.2   & -     & -     \\
ConvNeXt V2-T \cite{woo2023convnext}    & 224  & 28.6M  & 4.5  &  1127   & 82.5   & -     & -     \\
ViTAEv2-S+VSA \cite{zhang2022vsa}       & 224  & 19.8M  & 5.8  &  647    & 82.7   & 96.3  & -     \\
CSWin-T   \cite{dong2022cswin}          & 224  & 22.3M  & 4.3  &  1470   & 82.7   & 96.3  & 72.5  \\
HRViT-b3 \cite{gu2022multi}             & 224  & 37.9M  & 5.7  &  805    & 82.8   & -  & -  \\
DaViT-Tiny \cite{ding2022davit}         & 224  & 28.4M  & 4.5  &  1591   & 82.8   & 96.3  & 71.9  \\
ScalableViT-S \cite{yang2022scalablevit} & 224 & 32.0M  & 4.3  &  1379   & 83.1   & 96.3  & 72.4  \\
iFormer-S \cite{siinception}            & 224  & 19.9M  & 4.9  &  983    & 83.4   & 96.6  & 73.0  \\
Ortho-S \cite{huangorthogonal}          & 224  & 24M    & 4.5  &  -      & 83.4   & -     & -     \\
Dual-ViT-S  \cite{yao2022dual}          & 224  & 24.6M  & 5.4  &  1063   & 83.4   & 96.5  & 73.3  \\
CMT-S  \cite{guo2022cmt}                & 224  & 26.3M  & 4.1  &  848    & 83.5   & 96.6  & 73.4  \\
MaxViT-T  \cite{tu2022maxvit}           & 224  & 30.9M  & 5.6  &  788    & 83.6   & 96.6  & 73.0  \\
HIRI-ViT-S           & 224  & 34.8M  & 4.5  &  2028    & 83.6   & 96.7  & 73.2  \\
HIRI-ViT-S           & 384  & 34.8M  & 4.7  &  1833    & 84.0   & 96.8  & 73.7  \\
\multirow{2}{*}{HIRI-ViT-S}             & \multirow{2}{*}{448}  & \multirow{2}{*}{34.8M}  & \multirow{2}{*}{5.0}  &  \multirow{2}{*}{1490}   & \textbf{84.3} & \textbf{97.0} & \textbf{74.0}  \\
              &      &        &      &         & {\tiny ($\pm0.0278$)}       & {\tiny ($\pm0.0105$) }      & {\tiny ($\pm0.0293$)}  \\ \hline\hline
ResNet-101 \cite{he2016deep}            & 224  & 44.6M  & 7.9  &  2563   & 81.3  &  -    & -     \\
CvT-21 \cite{wu2021cvt}                 & 224  & 32.0M  & 7.1  &  1284   & 82.5   & -     & -     \\
CaiT-S24 \cite{touvron2021going}        & 224  & 46.9M  & 9.4  &  -      & 82.7   & -     & -     \\
ConvNeXt-S	\cite{liu2022convnet}       & 224  & 50.2M  & 8.7  &  1022   & 83.1  & 96.4  & 72.4     \\
Swin-S \cite{liu2021swin}               & 224  & 50.0M  & 8.7  &  1056   & 83.2   & 96.2  & 72.1  \\
Twins-SVT-B \cite{chu2021twins}         & 224  & 56.1M  & 8.6  &  1012   & 83.2   & 96.3  & 72.6  \\
PVTv2-B3 \cite{wang2021pvtv2}           & 224  & 45.2M  & 6.9  &  915    & 83.2   & 96.5  & 73.0  \\
SE-CoTNetD-101 \cite{li2022contextual}  & 224  & 40.9M  & 8.5  &  1075   & 83.2   & 96.5  & 73.0  \\
CoAtNet-1  \cite{dai2021coatnet}        & 224  & 42.2M  & 8.8  &  759    & 83.3   & -     & -     \\
Conformer-S \cite{peng2021conformer}    & 224  & 37.7M  & 10.6 &  653    & 83.4   & 96.5  & 73.6  \\
RegionViT-M+ \cite{chen2021regionvit}   & 224  & 42.0M  & 7.9  &  -      & 83.4   & -     & -     \\
PVTv2-B4 \cite{wang2021pvtv2}           & 224  & 62.6M  & 10.1 &  659    & 83.6   & 96.7  & 73.5  \\
CSWin-S  \cite{dong2022cswin}           & 224  & 34.6M  & 6.8  &  934    & 83.6   & 96.6  & 73.2  \\
Ortho-B \cite{huangorthogonal}          & 224  & 50M    & 8.6  &  -      & 84.0   & -     & -     \\
ScalableViT-B \cite{yang2022scalablevit} & 224 & 81.9M  & 8.8  &  1082   & 84.1   & 96.5  & 73.8  \\
DaViT-Small \cite{ding2022davit}        & 224  & 49.8M  & 8.8  &  1047   & 84.2   & 96.9  & 74.0  \\
Dual-ViT-B  \cite{yao2022dual}          & 224  & 42.1M  & 9.3  &  732    & 84.3   & 96.8  & 74.0  \\
MaxViT-S \cite{tu2022maxvit}            & 224  & 68.9M  & 11.7 &  556    & 84.5   & 96.8  & 73.9  \\
CMT-B  \cite{guo2022cmt}                & 256  & 45.7M  & 9.5  &  443    & 84.5   & 96.9  & 74.8  \\
iFormer-B \cite{siinception}            & 224  & 47.9M  & 9.4  &  593    & 84.6   & 97.0  & 74.6  \\
HIRI-ViT-B            & 224  & 54.4M  & 8.2  &  1187    & 84.7   & 97.0  & 74.8  \\
HIRI-ViT-B            & 384  & 54.4M  & 9.3  &  957    & 85.0   & 97.1  & 75.1  \\
\multirow{2}{*}{HIRI-ViT-B}      & \multirow{2}{*}{448}  & \multirow{2}{*}{54.4M}  & \multirow{2}{*}{9.9}  &  \multirow{2}{*}{880}    &  \textbf{85.2} &  \textbf{97.2}  & \textbf{75.2}  \\
              &      &        &      &         & {\tiny ($\pm0.0137$)}       & {\tiny ($\pm0.0134$) }      & {\tiny ($\pm0.0176$)} \\ \hline\hline
ConViT-B \cite{d2021convit}             & 224  & 86.5M  & 16.8 &  -      & 82.4   & 95.9  & -     \\
T2T-ViTt-24 \cite{yuan2021tokens}       & 224  & 64.1M  & 15.0 &  773    & 82.6   & 95.9  & 71.8  \\
HRFormer-B \cite{YuanFHLZCW21}          & 224  & 50.3M  & 13.7 &  302    & 82.8   & 96.3  & 71.9  \\
D-DW-Conv.-B  \cite{han2021connection}  & 224  & 161.6M & 13.0 &  1028   & 83.2   & 96.4  & 73.0  \\
CaiT-S36 \cite{touvron2021going}        & 224  & 68.4M  & 13.9 &  -      & 83.3   & -     & 72.5  \\
Swin-B \cite{liu2021swin}               & 224  & 88.0M  & 15.4 &  851    & 83.5   & 96.5  & 72.3  \\
Twins-SVT-L \cite{chu2021twins}         & 224  & 99.3M  & 15.1 &  817    & 83.7   & 96.5  & 73.4  \\
DeiT-B   \cite{touvron2021training}     & 224  & 86.6M  & 17.6 &  1151   & 83.8   & -     & -     \\
Focal-Base \cite{yang2021focal}         & 224  & 89.8M  & 16.0 &  241    & 83.8   & 96.5  & 73.4  \\
PVTv2-B5 \cite{wang2021pvtv2}           & 224  & 82.0M  & 11.8 &  640    & 83.8   & 96.6  & 73.4  \\
ViTAEv2-VSA-48M	\cite{zhang2022vsa}     & 224  & 51.1M  & 15.1 &  406    & 83.9   & 96.6  & -     \\
SE-CoTNetD-152 \cite{li2022contextual}  & 224  & 55.8M  & 17.0 &  732    & 84.0   & 97.0  & 73.8  \\
Conformer-B \cite{peng2021conformer}    & 224  & 83.3M  & 23.3 &  653    & 84.1   & 96.6  & 74.4  \\
Ortho-L \cite{huangorthogonal}          & 224  & 88M    & 15.4 &  -      & 84.2   & -     & -     \\
CSWin-B \cite{dong2022cswin}            & 224  & 77.4M  & 15.0 &  631    & 84.2   & 96.9  & 74.1  \\
ConvNeXt-L \cite{liu2022convnet}        & 224  & 197.8M & 34.4 &  513    & 84.3   & 96.9  & 74.2  \\
ScalableViT-L \cite{yang2022scalablevit} & 224 & 109.3M & 15.0 &  749    & 84.4   & 96.7  & 74.3  \\
ConvNeXt V2-L \cite{woo2023convnext}    & 224  & 198.0M & 34.4 &  371    & 84.5   & -     & -     \\
CoAtNet-3 \cite{dai2021coatnet}         & 224  & 167.0M & 36.7 &  366    & 84.5   & -     & -     \\
DaViT-Base \cite{ding2022davit}         & 224  & 88.0M  & 15.5 &  810    & 84.6   & 96.9  & 74.7  \\
ViTAEv2-B \cite{zhang2022vitaev2}       & 224  & 89.7M  & 24.3 &  446    & 84.6   & 96.9  & 74.4  \\
Dual-ViT-L \cite{yao2022dual}           & 224  & 72.5M  & 18.0 &  464    & 84.8   & 97.2  & 74.9  \\
iFormer-L \cite{siinception}            & 224  & 86.6M  & 14.1 &  491    & 84.8   & 97.0  & 74.9  \\
MaxViT-B \cite{tu2022maxvit}            & 224  & 119.5M & 23.4 &  318    & 85.0   & 97.0  & 74.5  \\
MaxViT-L \cite{tu2022maxvit}            & 224  & 211.8M & 43.9 &  241    & 85.2   & 97.0  & 75.1  \\
HIRI-ViT-L            & 224  & 94.4M & 17.0 &  660    & 85.3   & 97.2  & 75.2  \\
HIRI-ViT-L            & 384  & 94.4M & 18.2 &  544    & 85.5   & 97.4  & 75.7  \\
\multirow{2}{*}{HIRI-ViT-L}       & \multirow{2}{*}{448}  & \multirow{2}{*}{94.4M}  & \multirow{2}{*}{19.9} &  \multirow{2}{*}{488}    &  \textbf{85.7} &  \textbf{97.5} & \textbf{76.1} \\
              &      &        &      &         & {\tiny ($\pm0.0219$)}       & {\tiny ($\pm0.0068$) }      & {\tiny ($\pm0.0151$)} \\
\Xhline{2\arrayrulewidth}
\end{tabular}
\label{table:imagenet2}
\vspace{-0.2in}
\end{table}

\subsection{Normalization of Block}
ConvNets \cite{he2016deep,xie2017aggregated} usually use $\bf{BN}$ to stabilize the training process. $\bf{BN}$ can also be merged into convolution operation to speedup inference. In contrast, Vision Transformer backbones tend to normalize the features with layer normalization ($\bf{LN}$). $\bf{LN}$ is more friendly for dense prediction tasks with small training batch size (e.g., object detection and semantic segmentation) because it is independent of batch size. Compared to $\bf{BN}$, $\bf{LN}$ can also lead to slightly better performance \cite{liu2022convnet}. Nevertheless, $\bf{LN}$ results in heavier computational cost for high-resolution inputs. Accordingly, we utilize $\bf{BN}$ for the first three stages with high-resolution inputs, while $\bf{LN}$ is applied on the last two stages with low-resolution inputs. Moreover, we also replace $\bf{LN}$ with $\bf{BN}$ for $\bf{CFFN}$ block. By doing so, the inference procedure can speedup 7.6\%, while maintaining the performances.

\subsection{EMA Distillation}
During training, Exponential Moving Average (EMA) \cite{polyak1992acceleration} has been widely adopted to stabilize and improve the training procedure for both ConvNets \cite{tan2019efficientnet} and ViT \cite{touvron2021training}. Nevertheless, the message passing in conventional EMA is unidirectional, i.e., teacher network is updated by EMA based on the parameters of student network, thereby resulting in a sub-optimal solution. In an effort to trigger the bi-directional message interaction between teacher and student networks, we propose a new EMA distillation strategy to train HIRI-ViT. EMA distillation additionally leverages the probability distribution learnt from teacher network to guide the training of student network. In contrast to traditional knowledge distillation \cite{touvron2021training}, our EMA distillation does not rely on any extra large-scale pre-trained network.

Technically, given a pair of training samples ($x_a$, $y_a$) and ($x_b$, $y_b$), the new training sample ($\tilde x$, $\tilde y$) is generated by Cutmix/Mixup, which is further fed into student network $\mathcal{F}^s$ for network optimization. Taking Cutmix as an example, the new sample ($\tilde x$, $\tilde y$) can be generated as
\begin{equation}\small
\begin{aligned}
&{\tilde x = M \odot {x_a} + (1 - M) \odot {x_b}},\\
&{\tilde y = \lambda {y_a} + (1 - \lambda ){y_b},~\lambda = \frac{{\sum M}}{{HW}}},
\end{aligned}
\end{equation}
where $M \in {\{ 0,1\} ^{H \times W}}$ denotes the rectangular mask and $\odot$ is the element-wise multiplication. In EMA distillation, we feed the original samples $x_a$ and $x_b$ into teacher network $\mathcal{F}^t$ (the average pooling operation over the last feature map is removed), yielding the probability distribution map $P_a$ and $P_b$ respectively. Next, we assign a mixed target label for the generated sample $\tilde x$:
\begin{equation}\small
\begin{aligned}
{\tilde P = M \odot {P_a} + (1 - M) \odot {P_b}},~{\hat y = {\bf{AvgPool}}(\tilde P)},
\end{aligned}
\end{equation}
where $\bf{AvgPool}$ denotes the use of average pooling operation along spatial dimension. After that, we combine two target labels generated by Cutmix and teacher network:
${\rm{\bar y}} = {\alpha} {\tilde y} + (1-{\alpha}){\hat y}$,
where $\alpha$ is the tradeoff parameter. Finally, we feed the mixed sample ($\tilde x$, ${\rm{\bar y}}$) into student network $\mathcal{F}^s$ during training. In this way, the knowledge derived from teacher network is additionally exploited to facilitate the learning of student network.

\subsection{Architecture Details}
Table \ref{table:architecture} details the architectures of our HIRI-ViT family. Following the basic network configuration of existing CNN+ViT hybrid backbones \cite{liu2021swin,wang2021pyramid}, we construct three variants of our HIRI-ViT in different model sizes, i.e., HIRI-ViT-S (small size), HIRI-ViT-B (base size), and HIRI-ViT-L (large size). Specifically, the entire architecture of HIRI-ViT is composed of one HR-stem layer and five stages. For the first two stages with high-resolution inputs, we replace the conventional Transformer blocks with our light-weight High Resolution blocks to avoid huge computational overhead. For the third stage, we leverage only $\bf{CFFN}$ blocks to handle the middle-resolution feature maps. Similar to conventional Vision Transformer, we employ Transformer blocks in the last two stages with low-resolution inputs. For each stage $i$, $E_i$, $C_i$, and $HD_i$ represents the expansion ratio of feed-forward layer, channel dimension, and head number, respectively.

\section{Experiments}
We evaluate our HIRI-ViT on four vision tasks: image classification, object detection, instance segmentation, and semantic segmentation. In particular, HIRI-ViT is first trained from scratch on ImageNet-1K \cite{deng2009imagenet} for image classification task. Next, we fine-tune the pre-trained HIRI-ViT for the rest three downstream tasks: object detection and instance segmentation on COCO \cite{lin2014microsoft}, and semantic segmentation on ADE20K \cite{zhou2019semantic}.

\begin{table}[!tb] \small
\centering
\caption{Image classification performance comparison of HIRI-ViT with state-of-the-art regular backbones on the ImageNet-1K validation set at higher resolution. We group the backbones according to GFLOPs, and rank them based on Top-1 accuracy as in \cite{chen2022mobile}.}
\setlength{\tabcolsep}{2.5pt}
\begin{tabular}{c|c|c|c|cc}
\Xhline{2\arrayrulewidth}
Network       & Res. & Params & GFLOPs & Top-1 & Top-5 \\ \hline
CvT-13 \cite{wu2021cvt}       & 384 & 20.0M  & 16.3   & 83.0  & 96.4  \\
T2T-ViT-14 \cite{yuan2021tokens}   & 384 & 21.5M  & 17.1   & 83.3  & 96.5  \\
CeiT-S \cite{yuan2021incorporating}       & 384 & 24.2M  & 15.9   & 83.3  & 96.5  \\
ViT-S \cite{Touvron2022DeiTIR}        & 384 & 22.0M  & 15.5   & 83.4  & -     \\
CrossViT-15 \cite{chen2021crossvit}  & 384 & 28.5M  & 21.4   & 83.5  & -     \\
ViTAEv2-S \cite{zhang2022vitaev2}   & 384 & 19.2M  & 17.8   & 83.8  & 96.7  \\
iFormer-S \cite{siinception}   & 384 & 19.9M  & 16.1 & 84.6 & 97.3 \\
Ortho-S \cite{huangorthogonal} & 384 & 24M    & 14.3 & 84.8 & -    \\
HIRI-ViT-S                     & 768 & 34.8M  & 16.1   & \textbf{85.4}  & \textbf{97.5}  \\ \hline\hline
CvT-21 \cite{wu2021cvt}       & 384 & 32.0M  & 24.9   & 83.3  & 96.2  \\
CrossViT-18 \cite{chen2021crossvit}  & 384 & 44.6M  & 32.4   & 83.9  & -     \\
CaiT-XS36 \cite{touvron2021going}    & 384 & 38.6M  & 28.8   & 84.3  & -     \\
Ortho-B \cite{huangorthogonal} & 384  & 50M    & 26.6 & 85.2 & -    \\
iFormer-B \cite{siinception}   & 384  & 47.9M  & 30.5 & 85.7 & 97.6 \\
HIRI-ViT-B  & 768 & 54.4M  & 31.6     & \textbf{86.1}  & \textbf{97.7}  \\ \hline\hline
Swin-B \cite{liu2021swin}       & 384 & 88.0M  & 47.1   & 84.5  & 97.0  \\
Swin-B \cite{liu2021swin}       & 448 & 88.0M  & 61.9   & 84.9  & 97.1  \\
BoTNet-S1-128 \cite{srinivas2021bottleneck} & 384 & 75.1M  & 45.8   & 84.7  & 97.0  \\
ViT-B  \cite{Touvron2022DeiTIR}       & 384 & 86.9M  & 55.5   & 85.0  & -     \\
CaiT-S48 \cite{touvron2021going}     & 384 & 89.5M  & 63.8   & 85.1  & -     \\
ViTAEv2-B \cite{zhang2022vitaev2}    & 384 & 89.7M  & 74.4   & 85.3  & 97.1  \\
Ortho-L \cite{huangorthogonal} & 384  & 88M    & 47.4 & 85.4 & -    \\
iFormer-L \cite{siinception}   & 384  & 86.6M  & 45.3 & 85.8 & 97.6 \\
iFormer-L \cite{siinception}   & 448  & 88.4M  & 67.7 & 86.1 & 97.7 \\
HIRI-ViT-L  & 768 & 94.4M & 63.4   & \textbf{86.6}  & \textbf{97.8}  \\ \Xhline{2\arrayrulewidth}
\end{tabular}
\label{table:imagenethigh}
\end{table}

\subsection{Image Classification on ImageNet-1K}
\textbf{Setup.} ImageNet-1K dataset contains 1.28 million training images and 50,000 validation images over 1,000 object classes. During training, we adopt the common data augmentation strategies in \cite{liu2021swin,wang2021pvtv2}: random cropping, random horizontal flipping, Cutmix \cite{yun2019cutmix}, Mixup \cite{zhang2017mixup}, Random Erasing \cite{zhong2020random}, RandAugment \cite{cubuk2020randaugment}. The whole network is optimized via AdamW \cite{loshchilov2017decoupled} over 8 V100 GPUs, including 300 epochs with cosine decay learning rate scheduler \cite{loshchilov2016sgdr} and 5 epochs of linear warm-up on. The batch size, initial learning rate and weight decay are set as 1,024, 0.001 and 0.05, respectively. We report Top-1/5 accuracy on ImageNet-1K validation set, and Top-1 accuracy (i.e., V2 Top-1) on ImageNet V2 matched frequency test set \cite{recht2019imagenet} as in \cite{touvron2021training}.

\begin{table*}[!tb]\small
\centering
\caption{Comparison results of HIRI-ViT on COCO for object detection (basic detector: RetinaNet) and instance segmentation (basic model: Mask R-CNN). We split all runs into two groups according to GFLOPs (Small and Base). GFLOPs is calculated for baselines at 800 $\times$ 600 resolution as in \cite{chu2021twins}, while the input resolution of HIRI-ViT is twice larger (1,600 $\times$ 1,200).}
\setlength{\tabcolsep}{0.0pt}
\begin{tabular}{c|cc|cccccc|c|cc|cccccc}
\Xhline{2\arrayrulewidth}
\multicolumn{18}{c}{RetinaNet 1x \cite{lin2017focal}} \\ \hline
Backbone & Params & GFLOPs & $AP$ & $AP_{50}$ & $AP_{75}$ & $AP_S$ & $AP_M$ & $AP_L$ & Backbone & Params & GFLOPs & $AP$ & $AP_{50}$ & $AP_{75}$ & $AP_S$ & $AP_M$ & $AP_L$ \\ \hline
ResNet50 \cite{he2016deep}                   & 37.7M & 111 & 36.3 & 55.3 & 38.6 & 19.3 & 40.0 & 48.8    & ResNet101 \cite{he2016deep}                  & 56.7M & 149 & 38.5 & 57.8 & 41.2 & 21.4 & 42.6 & 51.1 \\
Swin-T   \cite{liu2021swin}                  & 38.5M & 118 & 41.5 & 62.1 & 44.2 & 25.1 & 44.9 & 55.5    & Swin-S \cite{liu2021swin}                    & 59.8M & 162 & 44.5 & 65.7 & 47.5 & 27.4 & 48.0 & 59.9  \\
Twins-SVT-S \cite{chu2021twins}              & 34.3M & 104 & 43.0 & 64.2 & 46.3 & 28.0 & 46.4 & 57.5    & RegionViT-B \cite{chen2021regionvit}         & -     &  -  & 44.6 & 66.4 & 47.6 & 29.6 & 47.6 & 59.0  \\
RegionViT-S+ \cite{chen2021regionvit}        & -     &   - & 43.9 & 65.5 & 47.3 & 28.5 & 47.3 & 57.9    & Twins-SVT-B \cite{chu2021twins}              & 67.0M & 163 & 45.3 & 66.7 & 48.1 & 28.5 & 48.9 & 60.6  \\
DaViT-Tiny \cite{ding2022davit}              & 38.5M & 117 & 44.0 & 65.6 & 47.3 & 29.6 & 47.9 & 57.3    & ScalableViT-B \cite{yang2022scalablevit}     & 85.6M & 171 & 45.8 & 67.3 & 49.2 & 29.9 & 49.5 & 61.0 \\
CMT-S \cite{guo2022cmt}                      & 44.3M & 126 & 44.3 & 65.5 & 47.5 & 27.1 & 48.3 & 59.1    & PVTv2-B3  \cite{wang2021pvtv2}               & 55.0M & 156 & 45.9 & 66.8 & 49.3 & 28.6 & 49.8 & 61.4  \\
PVTv2-B2  \cite{wang2021pvtv2}               & 35.1M & 122 & 44.6 & 65.6 & 47.6 & 27.4 & 48.8 & 58.6    & DaViT-Small \cite{ding2022davit}             & 59.9M & 160 & 46.0 & 67.8 & 49.0 & 30.2 & 50.2 & 59.9  \\
ScalableViT-S \cite{yang2022scalablevit}     & 36.4M & 123   & 45.2 & 66.5 & 48.4 & 29.2 & 49.1 & 60.3  & PVTv2-B5  \cite{wang2021pvtv2}               & 91.7M & 216 & 46.2 & 67.1 & 49.5 & 28.5 & 50.0 & 62.5 \\
HIRI-ViT-S                                 & 41.1M & 139 & \textbf{46.6} & \textbf{68.0} & \textbf{49.8} & \textbf{30.5} & \textbf{50.4} & \textbf{62.4} & HIRI-ViT-B & 59.8M & 202 & \textbf{48.4} & \textbf{69.7} & \textbf{51.9} & \textbf{33.4} & \textbf{52.6} & \textbf{63.4} \\ \hline\hline
\multicolumn{18}{c}{Mask R-CNN 1x \cite{he2017mask}} \\ \hline
Backbone & Params & GFLOPs & $AP^b$ & $AP^b_{50}$ & $AP^b_{75}$ & $AP^m$  & $AP^m_{50}$ & $AP^m_{75}$    & Backbone                                 & Params & GFLOPs & $AP^b$ & $AP^b_{50}$ & $AP^b_{75}$ & $AP^m$  & $AP^m_{50}$ & $AP^m_{75}$ \\ \hline
ResNet50 \cite{he2016deep}                   & 44.2M & 174 & 38.0 & 58.6  & 41.4  & 34.4 & 55.1  & 36.7  & ResNet101 \cite{he2016deep}              & 63.2M  & 210  & 40.4 & 61.1  & 44.2  & 40.4 & 61.1  & 44.2  \\
Swin-T   \cite{liu2021swin}                  & 47.8M & 177 & 42.2 & 64.6  & 46.2  & 39.1 & 61.6  & 42.0  & Swin-S \cite{liu2021swin}                & 69.1M  & 222  & 44.8 & 66.6  & 48.9  & 40.9 & 63.4  & 44.2  \\
Twins-SVT-S \cite{chu2021twins}              & 44.0M & 164 & 43.4 & 66.0  & 47.3  & 40.3 & 63.2  & 43.4  & Twins-SVT-B \cite{chu2021twins}          & 76.3M  & 224  & 45.2 & 67.6  & 49.3  & 41.5 & 64.5  & 44.8  \\
RegionViT-S+ \cite{chen2021regionvit}        & -     &  -  & 44.2 & 67.3  & 48.2  & 40.8 & 64.1  & 44.0  & RegionViT-B \cite{chen2021regionvit}     &  -     & -    & 45.4 & 68.4  & 49.6  & 41.6 & 65.2  & 44.8  \\
CMT-S \cite{guo2022cmt}                      & 44.5M & 170 & 44.6 & 66.8  & 48.9  & 40.7 & 63.9  & 43.4  & ScalableViT-B \cite{yang2022scalablevit} & 94.9M  & 215  & 46.8 & 68.7  & 51.5  & 42.5 & 65.8  & 45.9 \\
DaViT-Tiny \cite{ding2022davit}              & 47.9M & 161 & 45.0 & 68.1  & 49.4  & 41.1 & 64.9  & 44.2  & PVTv2-B3  \cite{wang2021pvtv2}           & 64.9M  & 200  & 47.0 & 68.1  & 51.7  & 42.5 & 65.7  & 45.7  \\
PVTv2-B2  \cite{wang2021pvtv2}               & 45.0M & 166 & 45.3 & 67.1  & 49.6  & 41.2 & 64.2  & 44.4  & PVTv2-B5  \cite{wang2021pvtv2}           & 101.6M & 260  & 47.4 & 68.6  & 51.9  & 42.5 & 65.7  & 46.0  \\
ScalableViT-S \cite{yang2022scalablevit}     & 46.3M & 167 & 45.8 & 67.6  & 50.0  & 41.7 & 64.7  & 44.8  & DaViT-Small \cite{ding2022davit}         & 69.3M  & 204  & 47.7 & 70.5  & 52.3  & 42.9 & 67.2  & 46.2 \\
iFormer-S \cite{siinception}	             & 39.6M & 187 & 46.2 & 68.5  & 50.6  & 41.9 & 65.3  & 45.0  & Ortho-B \cite{huangorthogonal}           & 69M	 & -	& 48.3 & 70.5  & 53.0  & 43.3 & 67.3  & 46.5 \\
Dual-ViT-S	\cite{yao2022dual}               & 41.9M & 178 & 46.5 & 68.3  & 51.2  & 42.2 & 65.3  & 46.1  & iFormer-B \cite{siinception}	            & 67.6M	 & 246	& 48.3 & 70.3  & 53.2  & 43.4 & 67.2  & 46.7 \\
Ortho-S	\cite{huangorthogonal}               & 44M	 & -   & 47.0 & 69.4  & 51.3  & 42.5 & 66.1  & 45.7  & Dual-ViT-B \cite{yao2022dual}	        & 58.6M	 & 233  & 48.4 & 69.9  & 53.3  & 43.4 & 66.7  & 46.8 \\
HIRI-ViT-S                                   & 51.0M & 183 & \textbf{47.3} & \textbf{69.4}  & \textbf{52.0}  & \textbf{43.0} & \textbf{66.3}  & \textbf{46.4} & HIRI-ViT-B  & 69.4M & 247 & \textbf{49.1} & \textbf{71.0}  & \textbf{54.3}  & \textbf{44.2} & \textbf{68.0}  & \textbf{47.8} \\
\Xhline{2\arrayrulewidth}
\end{tabular}
\label{table:od}
\vspace{-0.0in}
\end{table*}

\begin{table*}[!tb]\small
\centering
\caption{Comparison results of HIRI-ViT-S under four different object detectors on COCO for object detection. GFLOPs is calculated for baselines at 800 $\times$ 600 resolution as in \cite{chu2021twins}, while the input resolution of HIRI-ViT is twice larger (1,600 $\times$ 1,200).}
\begin{tabular}{c|cc|ccc|c|cc|ccc}
\Xhline{2\arrayrulewidth}
Backbone                                  & Params & GFLOPs      & $AP^b$ & $AP^b_{50}$ & $AP^b_{75}$    & Backbone & Params & GFLOPs   & $AP^b$ & $AP^b_{50}$ & $AP^b_{75}$ \\ \hline
\multicolumn{6}{c|}{Cascade Mask R-CNN} & \multicolumn{6}{c}{ATSS}    \\ \hline
ResNet50 \cite{he2016deep}                & 82.0M  & 637 & 46.3 & 64.3  & 50.5                           & ResNet50 \cite{he2016deep}    & 32.1M  & 97  & 43.5 & 61.9  & 47.0    \\
Swin-T  \cite{liu2021swin}                & 85.6M  & 639 & 50.5 & 69.3  & 54.9                           & Swin-T  \cite{liu2021swin}    & 35.7M  & 100 & 47.2 & 66.5  & 51.3    \\
PVTv2-B2 \cite{wang2021pvtv2}             & 82.9M  & 644 & 51.1 & 69.8  & 55.3                           & PVTv2-B2 \cite{wang2021pvtv2} & 33.0M  & 106  & 49.9 & 69.1  & 54.1    \\
Dual-ViT-S \cite{yao2022dual}           & 79.7M  & 657 & 52.4 & 71.0 & 56.9                              & Dual-ViT-S \cite{yao2022dual} & 29.8M & 118 & 51.0 &  69.9 &  \textbf{55.9}  \\
HIRI-ViT-S                              & 88.9M  & 661 & \textbf{52.6} & \textbf{71.3} & \textbf{57.1} & HIRI-ViT-S   & 39.0M  & 123 & \textbf{51.5} & \textbf{71.1} & \textbf{55.9} \\ \hline\hline
\multicolumn{6}{c|}{GFL}  & \multicolumn{6}{c}{Sparse RCNN}                      \\ \hline
ResNet50 \cite{he2016deep}                & 32.2M  & 98  & 44.5 & 63.0  & 48.3                           & ResNet50 \cite{he2016deep}    & 106.1M & 75 & 44.5 & 63.4  & 48.2    \\
Swin-T  \cite{liu2021swin}                & 35.9M  & 102 & 47.6 & 66.8  & 51.7                           & Swin-T \cite{liu2021swin}     & 109.7M & 93 & 47.9 & 67.3  & 52.3    \\
PVTv2-B2 \cite{wang2021pvtv2}             & 33.1M  & 108  & 50.2 & 69.4  & 54.7                           & PVTv2-B2 \cite{wang2021pvtv2} & 107.0M & 102 & 50.1 & 69.5  & 54.9    \\
Dual-ViT-S \cite{yao2022dual}           & 30.0M &  120 &  51.3 &  70.1 &  55.7                           & Dual-ViT-S \cite{yao2022dual}  & 103.8M &  110 &  51.4 &  70.9 &  56.2  \\
HIRI-ViT-S                              & 39.1M  & 125 & \textbf{51.8} & \textbf{71.0} & \textbf{56.2} & HIRI-ViT-S & 113.0M & 119  & \textbf{51.9} & \textbf{71.9} & \textbf{56.6} \\ \Xhline{2\arrayrulewidth}
\end{tabular}
\label{table:od4}
\end{table*}

\textbf{Performance Comparison.}
Table \ref{table:imagenet2} shows the performance comparisons between our HIRI-ViT family and existing CNN/ViT backbones. It is worthy to note that all the baselines are fed with typical resolution inputs (224$\times$224), while our HIRI-ViT family scales up Vision Transformer with high resolution inputs (448$\times$448). Overall, under comparable computational cost in each group, our HIRI-ViT (448$\times$448) achieves consistent performance improvements against state-of-the-art backbones across all model sizes. Remarkably, for large model size (backbones with GFLOPs more than 11.7), the Top-1 accuracy of our HIRI-ViT-L (448$\times$448) is 85.7\%, which leads to the absolute performance gain of 0.5\% than the best competitor MaxViT-L (85.2\%). Although the input resolution of HIRI-ViT-L is significantly larger than MaxViT-L, our HIRI-ViT-L (448$\times$448) requires less GFLOPs than MaxViT-L, and indicates the advantage of faster inference speed with almost doubled throughput. Such results clearly demonstrate that our HIRI-ViT achieves better balance between performance and computation cost, especially tailored for high resolution inputs. It is also worthy to note that when feeding with the typical resolution inputs (224$\times$224), our HIRI-ViT under each model size achieves comparable performances in comparison to state-of-the-art backbones, while requiring significantly less computational cost. For example, in the group of large model size, the Top-1 accuracy of our HIRI-ViT-L (224$\times$224) is 85.3\% and the corresponding throughput is 660 images per second, which is extremely faster than the best competitor MaxViT-L (Top-1 accuracy: 85.2\%, Throughput: 241 images per second). The results again confirm the cost-efficient design of our HIRI-ViT.

\textbf{Performance Comparison at Higher Resolution.}
Table \ref{table:imagenethigh} illustrates the comparisons between our HIRI-ViT family and other state-of-the-art vision backbones with larger input image size. For this upgraded HIRI-ViT with higher resolution inputs (768 $\times$ 768), we adopt the AdamW optimizer \cite{loshchilov2017decoupled} on 8 V100 GPUs, with a momentum of 0.9, an initial learning rate of $1.0e^{-5}$, and the weight decay of $1.0e^{-8}$. The optimization process includes 30 epochs with cosine decay learning rate scheduler \cite{loshchilov2016sgdr}. Similarly, for each group with comparable computational cost, our HIRI-ViT consistently obtains performance gains in comparison to other vision backbones at higher resolution. These results clearly validate the effectiveness of our proposed five-stage ViT backbone tailored for high-resolution inputs. This design novelly decomposes the typical CNN operations into both high-resolution and low-resolution branches in parallel, and thus maintains favorable computational cost even under the setup with higher-resolution inputs.

\subsection{Object Detection and Instance Segmentation on COCO}
\textbf{Setup.} We conduct both object detection and instance segmentation tasks on COCO dataset. We adopt the standard setting in \cite{liu2021swin,wang2021pvtv2} and train all models on COCO-2017 training set ($\sim$118K images). The learnt model is finally evaluated over COCO-2017 validation set (5K images). We use two mainstream detectors (RetinaNet \cite{lin2017focal} and Mask R-CNN \cite{he2017mask}) for object detection and instance segmentation. The primary CNN backbones in each detector are replaced with our HIRI-ViT family (initially pre-trained on ImageNet-1K). All the other newly added layers are initialized with Xavier \cite{glorot2010understanding}. We fine-tune the detectors on 8 V100 GPUs via AdamW optimizer \cite{loshchilov2017decoupled} (batch size: 16). For RetinaNet and Mask R-CNN, we adopt the standard 1$\times$ training schedule (12 epochs). The shorter side of each image is resized to 1,600 pixels, while the longer side does not exceed 2,666 pixels. For object detection, we conduct experiments on four additional object detection
methods: Cascade Mask R-CNN \cite{cai2018cascade}, ATSS \cite{zhang2020bridging}, GFL \cite{li2020generalized}, and Sparse RCNN \cite{sun2021sparse}. Following \cite{liu2021swin,wang2021pvtv2}, the 3$\times$ schedule (36 epochs) with multi-scale strategy is utilized for training. The input image is randomly resized by maintaining the shorter side within the range of [960, 1,600], while the longer side is forced to be less than 2,666 pixels. We report the Average Precision score ($AP$) across different IoU thresholds and three different object sizes, i.e., small ($AP_S$), medium ($AP_M$), large ($AP_L$). For instance segmentation task, the bounding box and mask AP scores ($AP^b$, $AP^m$) are reported.

\begin{table}[!tb]\small
\setlength{\tabcolsep}{1pt}
\centering
\caption{Comparison results of HIRI-ViT for Semantic segmentation on ADE20K dataset. All runs are split into two groups w.r.t the GFLOPs (Small and Base). We calculate GFLOPs at the resolution of 512 $\times$ 512 as in \cite{chu2021twins}.}
\begin{tabular}{cc|cc|c}
\Xhline{2\arrayrulewidth}
Method       & Backbone   & Params  & GFLOPs                     & mIoU    \\ \hline
UPerNet \cite{xiao2018unified}           & ResNet-50 \cite{he2016deep}          & 66.5M & 238 & 42.8  \\
DeeplabV3 \cite{chen2018encoder}         & ResNeSt-50 \cite{zhang2020resnest}   &  -    &  -  & 45.1  \\
Semantic FPN \cite{kirillov2019panoptic} & PVTv2-B2 \cite{wang2021pvtv2}  &  -    &  -  & 45.2  \\
UPerNet \cite{xiao2018unified}           & DeiT-S \cite{touvron2021training}    &  -    &  -  & 45.6  \\
UPerNet \cite{xiao2018unified}           & Swin-T \cite{liu2021swin}            & 59.9M & 237 & 45.8  \\
UPerNet \cite{xiao2018unified}           & DaViT-Tiny \cite{ding2022davit}      & 60.0M & 236 & 46.3  \\
UPerNet \cite{xiao2018unified}           & Twins-SVT-S \cite{chu2021twins}      & 54.4M & 228 & 47.1  \\
UPerNet \cite{xiao2018unified}           & Ortho-S \cite{huangorthogonal}       & 54M   & -   & 48.5  \\
UPerNet \cite{xiao2018unified}           & ScalableViT-S \cite{yang2022scalablevit}  & 56.1M   & 235   & 48.5  \\
UPerNet \cite{xiao2018unified}           & HRViT-b2 \cite{gu2022multi} & 49.7M & 233	& 49.1 \\
UPerNet \cite{xiao2018unified}           & HIRI-ViT-S          & 61.4M & 240 & \textbf{50.3} \\ \hline\hline
UPerNet \cite{xiao2018unified}           & ResNet-101 \cite{he2016deep}          & 85.5M & 257    & 44.9 \\
DeeplabV3 \cite{chen2018encoder}         & ResNeSt-101 \cite{zhang2020resnest}   &  -    &  -     & 46.9 \\
Semantic FPN \cite{kirillov2019panoptic} & PVTv2-B3 \cite{wang2021pvtv2}         &  -    &  -     & 47.3 \\
UPerNet \cite{xiao2018unified}           & DaViT-Small \cite{ding2022davit}      & 81.4M & 260      & 48.8 \\
UPerNet \cite{xiao2018unified}           & Twins-SVT-B \cite{chu2021twins}       & 88.5M & 261    & 48.9 \\
UPerNet \cite{xiao2018unified}           & DeiT-B  \cite{touvron2021training}    &  -    &  -     & 49.3 \\
UPerNet \cite{xiao2018unified}           & Swin-S \cite{liu2021swin}             & 81.3M & 261    & 49.5 \\
UPerNet \cite{xiao2018unified}           & ScalableViT-B \cite{yang2022scalablevit}    & 107.0M  & 261      & 49.5 \\
UPerNet \cite{xiao2018unified}           & Ortho-B \cite{huangorthogonal}        & 81M   & -      & 49.8 \\
OCR	    \cite{yuan2020object}            & HRFormer-B \cite{YuanFHLZCW21}        & 56.2M & - & 50.0 \\
UPerNet \cite{xiao2018unified}           & ConvNeXt V2-B \cite{woo2023convnext}  & 122.2M  & 293      & 50.5 \\
UPerNet \cite{xiao2018unified}           &  HIRI-ViT-B  & 80.6M & 265 & \textbf{51.8}  \\ \Xhline{2\arrayrulewidth}
\end{tabular}
\label{table:ss}
\end{table}

\textbf{Performance Comparison.} Table \ref{table:od} summarizes the object detection and instance segmentation performances of RetinaNet and Mask R-CNN with different backbones on COCO benchmark. Our HIRI-ViT-S and HIRI-ViT-B manage to consistently exhibit better performances across all metrics than other backbones in each group with comparable computational costs. Concretely, HIRI-ViT-S outperforms the best competitor ScalableViT-S by 1.4\% ($AP$) and Ortho-S by 0.5\% ($AP^m$) on the basis of RetinaNet and Mask R-CNN detector, respectively. Meanwhile, the input resolution of HIRI-ViT-S is twice larger than that of ScalableViT-S, while both of them require similar parameter number and GFLOPs. This clearly validates the superior generalizability of scaling up Vision Transformer via our design in downstream tasks. Table \ref{table:od4} further shows the performances of four additional object detectors under different backbones on COCO. Similarly, HIRI-ViT-S leads to consistent performance gains against other baselines for each object~detector.

\subsection{Semantic Segmentation on ADE20K}
\textbf{Setup.} We next evaluate HIRI-ViT for semantic segmentation on ADE20K. This dataset covers 150 semantic categories and contains 20,000 images for training, 2,000 for validation, and 3,000 for testing. Here we follow \cite{liu2021swin} and use UPerNet \cite{xiao2018unified} as the base model, where the CNN backbone is replaced with our HIRI-ViT. The whole network is trained with 160K iterations over 8 V100 GPUs via AdamW optimizer \cite{loshchilov2017decoupled}. We utilize the linear learning rate decay scheduler with 1,500 iterations linear warmup for optimization. The batch size, weight decay, and initial learning rate are set as 16, 0.01, and 0.00006. We adopt the standard data augmentations: random horizontal flipping, random photometric distortion, and random re-scaling within the ratio range of [0.5, 2.0]. All the other hyperparameters and detection heads are set as in Swin \cite{liu2021swin} for fair comparison.

\textbf{Performance Comparison.}
Table \ref{table:ss} details the performances of different backbones on ADE20K validation set for semantic segmentation. Similar to the observations in object detection and instance segmentation downstream tasks, our HIRI-ViT-S and HIRI-ViT-B attain the best mIoU score within each group with comparable computational costs. Specifically, under the same base model of UPerNet, HIRI-ViT-S boosts up the mIoU score of HRViT-b2 by 1.2\%, which again demonstrates the effectiveness of our proposal.

\begin{table}[!tb]\small
\setlength{\tabcolsep}{0.0pt}
\centering
\caption{Ablation study by progressively considering each design in HIRI-ViT-S on ImageNet-1K for image classification.}
\begin{tabular}{c|c|l|c|c|cc}
\Xhline{2\arrayrulewidth}
Row & \#stage & \multicolumn{1}{c|}{Model}       & Params & GFLOPs & Top-1 & Top-5 \\ \hline
1   & 4 & M-ViT                            & 35.0M  &  4.4   & 82.7  &  96.3 \\
2   & 4 & FFN $\rightarrow$ CFFN           & 35.1M  &  4.4   & 82.9  &  96.4 \\
3   & 4 & Remove MHA                       & 34.2M  &  4.3   & 82.9  &  96.4 \\
4   & 4 & LN $\rightarrow$ BN              & 34.2M  &  4.2   & 83.0  &  96.4 \\
5   & 4 & ViT-stem $\rightarrow$ Conv-stem & 34.2M  &  4.5   & 83.4  &  96.7 \\
6   & 4 & IRDS                             & 34.5M  &  4.7   & 83.6  &  96.8 \\
7   & 4 & Resolution 224 $\rightarrow$ 448 & 34.5M  &  21.5  & 84.6  &  97.3 \\ \hline\hline
8   & 5 & Four Stages$\rightarrow$Five Stages & 34.7M  &  5.8   & 84.0  &  96.9 \\
9   & 5 & Conv-stem $\rightarrow$ HR-stem  & 34.7M  &  5.1   & 84.0  &  96.8 \\
10  & 5 & HR block                         & 34.8M  &  5.0   & 84.1  &  96.9 \\
11  & 5 & EMA Distillation               & 34.8M  &  5.0   & 84.3  &  97.0 \\ \Xhline{2\arrayrulewidth}
\end{tabular}
\label{table:ab}
\end{table}

\subsection{Ablation Study}
In this section, we first illustrate how to construct a strong multi-stage Vision Transformer (M-ViT) with four stages, which acts as a base model. Then, we extend the base model with five stages, and study how each design in HIRI-ViT influences the overall performances on ImageNet-1K for image classification task. Table \ref{table:ab} summarizes the performances of different ablated runs by progressively considering each design into the base model.

\textbf{M-ViT.} We start from a base model, i.e., multi-stage Vision Transformer (M-ViT) with four stages. The whole architecture of M-ViT is similar to PVT \cite{wang2021pyramid}, which consists of one ViT-stem layer and four stages. Each stage contains a stack of Transformer blocks. Each Transformer block is composed of $\bf{MHA}$ and $\bf{FFN}$. In the first two stages, M-ViT utilizes a strided convolution as spatial reduction to downsample the keys and values, while the last two stages do not use any spatial reduction operation. A single strided convolution is leveraged to perform spatial downsampling and meanwhile enlarge the channel dimension between every two stages. $\bf{LN}$ is adopted for feature normalization. As indicated in Table \ref{table:ab} (Row 1), the top-1 accuracy of M-ViT achieves 82.7\%.

\textbf{FFN$\rightarrow$CFFN.} Row 2 upgrades the base model by integrating each $\bf{FFN}$ with additional depth-wise convolution (i.e., $\bf{CFFN}$). In this way, $\bf{CFFN}$ enables the exploitation of inductive bias and thus boosts up performance (82.9\%).

\begin{figure*}[!tb]
\centering {\includegraphics[width=0.95\linewidth]{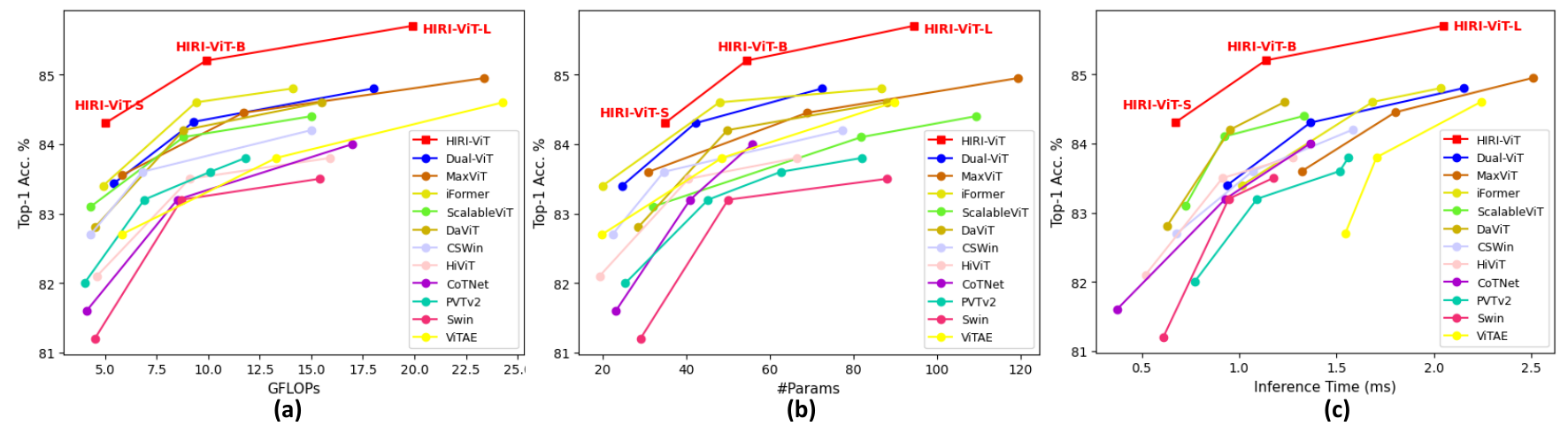}}
\vspace{-0.15in}
\caption{Computational Cost (i.e., GFLOPs, model parameter number, and inference time) vs. accuracy on ImageNet-1K.}
\vspace{-0.2in}
\label{fig:curve}
\end{figure*}

\textbf{Remove MHA.} Next, we take the inspiration from \cite{dai2021coatnet}, and remove the multi-head self-attention ($\bf{MHA}$) in the first two stages. The channel expansion dimension in $\bf{CFFN}$ is also enlarged. As shown in Table \ref{table:ab} (Row 3), this ablated run maintains the performances, while both GFLOPs and parameter number decrease.

\textbf{LN$\rightarrow$BN.} Then, we replace $\bf{LN}$ with $\bf{BN}$ in the first two stages and each $\bf{CFFN}$ block (Row 4). The top-1 accuracy slightly increases to 83.0\%, showing that $\bf{BN}$ is more suitable than $\bf{LN}$ for blocks with convolution operations.

\textbf{ViT-stem$\rightarrow$Conv-stem.} When we replace ViT-stem with Conv-stem (Row 5), the top-1 accuracy is further improved to 83.4\%. This observation demonstrates the merit of Conv-stem that injects a small dose of inductive bias in the early visual processing, thereby stabilizing the optimization and improving peak performance.

\textbf{IRDS.} After that, we apply the inverted residual downsampling between every two stages (Row 6). The final version of M-ViT with four stages manages to achieve the top-1 accuracy of 83.6\% , which is competitive and even surpasses most existing ViT backbones.

\textbf{Resolution 224$\rightarrow$448.} Next, we directly enlarge the input resolution from 224$\times$224 to 448$\times$448 for this four-stage structure (Row 7). This ablated run leads to clear performance boosts, while sharply increasing GFLOPs from 4.7 to 21.5. This observation aligns with existing four-stage ViT architectures (e.g., PVT and Swin Transformer) where the computational cost scales quadratically for enlarged input resolution.

\textbf{Four Stages$\rightarrow$Five Stages.}
Furthermore, we extend the four-stage ablated run (Row 7) by leveraging additional stage to further downsample high-resolution inputs. Such five-stage structure (Row 8) significantly decreases GFLOPs with high-resolution inputs (448$\times$448) and the Top-1 accuracy drops into 84.0\%, which still outperforms another four-stage ablated run (Row 6) under 224$\times$224 inputs. The results basically confirm the effectiveness of our five-stage structure that seeks better cost-performance balance for high-resolution inputs.

\textbf{Conv-stem$\rightarrow$HR-stem.} Then we replace Conv-stem with our HR-stem (remoulded in two-branch design). As shown in Table \ref{table:ab} (Row 9), the GFLOPs is clearly dropped (5.1), but the top-1 accuracy still maintains (84.0\%).

\textbf{HR block.} After that, we replace each $\bf{CFFN}$ with two stacked HR blocks in the first two stages (Row 10). Note that the Top-1 accuracy on ImageNet are almost saturated ($\sim$84.0\%) for small size (GFLOPs: $\sim$5.0) and it is relatively difficult to introduce large margin of improvement. However, our HR block still manages to achieve a performance gain of 0.1\% in both Top-1 and Top-5 accuracies and meanwhile slightly reduces computational cost, which again validates the advantage of the cost-efficient encoding via our high and low-resolution branches in parallel.

\textbf{EMA Distillation.} Finally, Row 11 is the full version of our HIRI-ViT-S that is optimized with additional EMA distillation, leading to the best top-1 accuracy (84.3\%).

\subsection{Impact of Each Kind of Block in Five Stages}

Here we further conduct additional ablation study to fully examine the impact of each kind of block (HR/CFFN/Transformer block) in five stages. Table \ref{table:ab2} shows the performances of different ablated variants of HIRI-ViT by constructing five stages with different blocks. Concretely, the results of Row 1-3 indicate that replacing more HR blocks with Transformer blocks in the last three stages can generally lead to performance improvement, but suffer from heavier computational cost. For example, the use of Transformer block in the third stage (Row 3) obtains marginal performance gain (0.1\% in Top-1 accuracy), while GFLOPs is clearly increased. The results basically validate the effectiveness of Transformer block with low-resolution inputs in the late stages, but Transformer block naturally results in huge computational overhead especially for high-resolution inputs in the early stages. Furthermore, we replace Transformer block with CFFN block to handle middle-resolution inputs in the third stage (Row 4). Such design nicely reduces GFLOPs and meanwhile maintains the Top-1 and Top-5 accuracies, which demonstrates the best trade-off between computational cost and performance in HIRI-ViT.

\begin{table}[!tb]\small
\vspace{-0.25in}
\setlength{\tabcolsep}{0.5pt}
\centering
\caption{Ablation study by constructing five stages with different combinations of HR block (HR), CFFN block (CFFN), and Transformer block (T) in HIRI-ViT-S on ImageNet-1K for image classification.}
\begin{tabular}{l|c|c|cc}
\Xhline{2\arrayrulewidth}
\multicolumn{1}{c|}{Model}       & Params & GFLOPs & Top-1 & Top-5 \\ \hline
HR $\rightarrow$ HR $\rightarrow$ HR $\rightarrow$ HR $\rightarrow$ T    & 33.2M  &  4.3   & 83.0  &  96.5 \\
HR $\rightarrow$ HR $\rightarrow$ HR $\rightarrow$ T $\rightarrow$ T     & 34.7M  &  4.8   & 84.2  &  96.9 \\
HR $\rightarrow$ HR $\rightarrow$ T $\rightarrow$ T $\rightarrow$ T      & 34.8M  &  5.3   & 84.3  &  97.0 \\
HR $\rightarrow$ HR $\rightarrow$ CFFN $\rightarrow$ T $\rightarrow$ T   & 34.8M  &  5.0   & 84.3  &  97.0 \\ \Xhline{2\arrayrulewidth}
\end{tabular}
\vspace{-0.25in}
\label{table:ab2}
\end{table}

\subsection{Computational Cost vs. Accuracy}
Figure \ref{fig:curve} further illustrates the accuracy curves with regard to computational cost (i.e., (a) GFLOPs, (b) model parameter number and (c) inference time) for our HIRI-ViT and other state-of-the-art vision backbones. As shown in this figure, the curves of our HIRI-ViT backbones are always over the ones of other vision backbones. That is, our HIRI-ViT backbones seek better computational cost-accuracy tradeoffs than existing vision backbones.

\subsection{Extension to Other Backbones}
Here we report the performance/computational cost by leveraging our proposal to scale up three different vision backbones (HRNet, PVTv2, DaViT) with higher-resolution inputs (224$\times$224 to 448$\times$448). As shown in Table \ref{table:ab3}, our five-stage structure with two-branch design in HIRI-ViT consistently leads to significant performance boost for each vision backbone, while retaining favorable computational cost. The results further validate the generalizability of our five-stage structure with two-branch design for scaling up vision backbones with high-resolution inputs.

\begin{table}[!tb]\small
\setlength{\tabcolsep}{0.5pt}
\centering
\caption{Effect of applying HIRI-ViT into different vision backbones (HRNet, PVTv2, DaViT) for scaling up with higher-resolution inputs (224$\times$224 to 448$\times$448) on ImageNet-1K for image classification.}
\begin{tabular}{l|c|c|c|cc}
\Xhline{2\arrayrulewidth}
Network       & Res. & Params & GFLOPs & Top-1 & Top-5 \\ \hline
HRNet-W18  \cite{wang2020deep}  & 224 & 21.3M & 4.3  & 79.9 & 94.9 \\
HRNet-W18  \cite{wang2020deep} (+ HIRI-ViT) & 448 & 21.4M & 4.6  & 80.5 & 95.2 \\\hline\hline
PVTv2-B2   \cite{wang2021pvtv2} & 224 & 25.4M & 4.0  & 82.0 & 95.6 \\
PVTv2-B2   \cite{wang2021pvtv2} (+ HIRI-ViT) & 448 & 25.5M & 4.4  & 82.6 & 96.2 \\\hline\hline
DaViT-Tiny \cite{ding2022davit} & 224 & 28.4M & 4.5  & 82.8 & 96.3 \\
DaViT-Tiny \cite{ding2022davit} (+ HIRI-ViT) & 448 & 28.5M & 5.0  & 83.3 & 96.5 \\
\Xhline{2\arrayrulewidth}
\end{tabular}
\label{table:ab3}
\end{table}

\section{Conclusions}

In this work, we design a new five-stage ViT backbone for high-resolution inputs, namely HIRI-ViT, that novelly decomposes the typical CNN operations into both high-resolution and low-resolution branches in parallel. With such principled five-stage and two-branch design, our HIRI-ViT is armed with ability to scale up Vision Transformer backbone with high-resolution inputs in a cost-efficient fashion. Extensive experiments are conducted on ImageNet-1K (image classification), COCO (object detection and instance segmentation) and ADE20K datasets (semantic segmentation) to validate the effectiveness of our HIRI-ViT against competitive CNN or ViT backbones.

In spite of these observations, open problems remain. While our five-stage structure with two-branch design has clearly improved the efficiency for scaling up Vision Transformer, we observe less improvement on performance/computational cost when employing six-stage structure. Moreover, how to scale up Video Vision Transformer with high-resolution inputs still present a major challenge.


%

%
%
%
%
%
%
%
%
%
%
%


\begin{IEEEbiography}[{\includegraphics[width=1in,height=1.25in,clip]{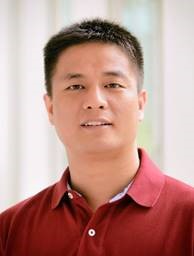}}]{Ting Yao}
is currently the CTO of HiDream.ai, a high-tech startup company focusing on generative intelligence for creativity. Previously, he was a Principal Researcher with JD AI Research in Beijing, China and a Researcher with Microsoft Research Asia in Beijing, China. Dr. Yao has co-authored more than 100 peer-reviewed papers in top-notch conferences/journals. He has developed one standard 3D Convolutional Neural Network, i.e., Pseudo-3D Residual Net, for video understanding, and his video-to-text dataset of MSR-VTT has been used by 400+ institutes worldwide. He serves as an associate editor of IEEE Transactions on Multimedia, Pattern Recognition Letters, and Multimedia Systems. His works have led to many awards, including 2015 ACM-SIGMM Outstanding Ph.D. Thesis Award, 2019 ACM-SIGMM Rising Star Award, 2019 IEEE-TCMC Rising Star Award, 2022 IEEE ICME Multimedia Star Innovator Award, and the winning of 10+ championship in worldwide competitions.
\end{IEEEbiography}

\begin{IEEEbiography}[{\includegraphics[width=1in,height=1.25in,clip]{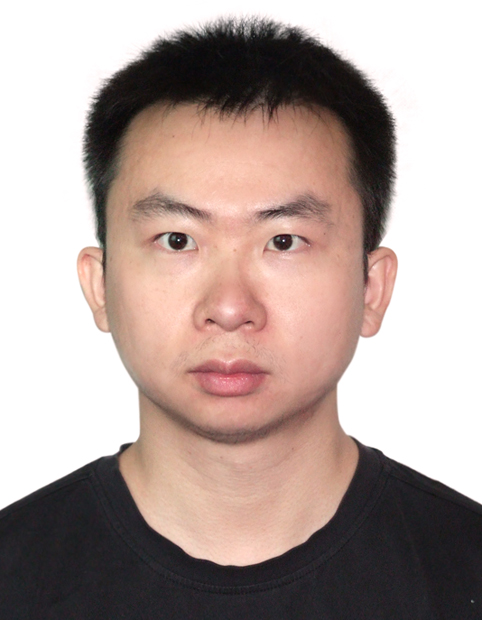}}]{Yehao Li}
is currently a Senior Researcher in HiDream.ai. He received Ph.D. degree in 2019 from Sun Yet-sen University (SYSU), Guangzhou, China. He was one of core designers of top-performing multimedia analytic systems in worldwide competitions such as COCO Image Captioning, ActivityNet Dense-Captioning Events in Videos Challenge 2017 and Visual Domain Adaptation Challenge 2018. His research interests include vision and language, large-scale visual search, and video understanding.
\end{IEEEbiography}

\begin{IEEEbiography}[{\includegraphics[width=1in,height=1.25in,clip]{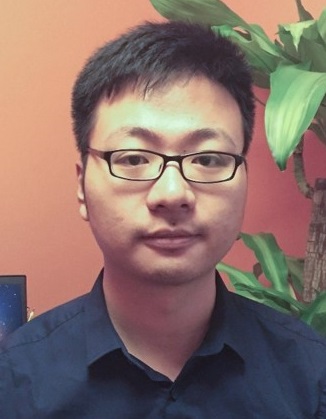}}]{Yingwei Pan}
is currently a Technical Director in HiDream.ai. His research interests include vision and language, and visual content understanding. He is the principal designer of the top-performing multimedia analytic systems in international competitions such as COCO Image Captioning, Visual Domain Adaptation Challenge 2018, ActivityNet Dense-Captioning Events in Videos Challenge 2017, and MSR-Bing Image Retrieval Challenge 2014 and 2013. He received Ph.D. degree in Electrical Engineering from University of Science and Technology of China in 2018.
\end{IEEEbiography}

\begin{IEEEbiography}[{\includegraphics[width=1in,height=1.25in,clip,keepaspectratio]{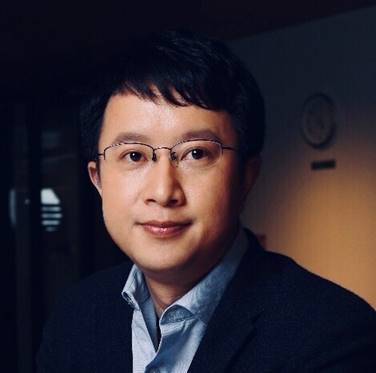}}]{Tao Mei}
(Fellow, IEEE) is the Founder and CEO of HiDream.ai. Previously, He was a Vice President of JD.COM and a Senior Research Manager of Microsoft Research. He has authored or co-authored over 200 publications (with 12 best paper awards) in journals and conferences, 10 book chapters, and edited five books. He holds over 25 US and international patents. He is or has been an Editorial Board Member of IEEE Trans. on Image Processing, IEEE Trans. on Circuits and Systems for Video Technology, IEEE Trans. on Multimedia, ACM Trans. on Multimedia Computing, Communications, and Applications, Pattern Recognition, etc. He is the General Co-chair of IEEE ICME 2019, the Program Co-chair of ACM Multimedia 2018, IEEE ICME 2015 and IEEE MMSP 2015.
	
Tao received B.E. and Ph.D. degrees from the University of Science and Technology of China, Hefei, China, in 2001 and 2006, respectively. He is a Fellow of IEEE (2019), a Fellow of IAPR (2016), a Distinguished Scientist of ACM (2016), and a Distinguished Industry Speaker of IEEE Signal Processing Society (2017).

\end{IEEEbiography}

\end{document}